\documentclass[runningheads]{llncs}
\usepackage[T1]{fontenc}
\usepackage{graphicx}
\usepackage{amsmath}
\usepackage{enumitem}
\usepackage{amssymb}  % common math symbols
\usepackage{bm}  
\usepackage{booktabs}
\usepackage{array}
\usepackage{adjustbox}
\usepackage{multirow}  
\usepackage{algorithm}
\usepackage{algorithmic}

\usepackage{hyperref}
\hypersetup{
	colorlinks=true,
	linkcolor=black,
	filecolor=magenta,
	urlcolor=cyan,
	citecolor=blue,
}
\usepackage{color}

\begin{document}
\title{\textit{DARD}: Dice Adversarial Robustness Distillation against Adversarial Attacks}
\titlerunning{DARD: Dice Adversarial Robustness Distillation}
\author{Jing Zou\(^1\),
Shungeng Zhang\(^1\),
Meikang Qiu\(^1\)\(^*\)
Chong Li\(^2\)}
\authorrunning{J. Zou et al.}
\institute{%
\textsuperscript{1} 
Augusta University, Augusta, GA 30901, USA\\%
\textsuperscript{2} Electrical Engineering Dept., Columbia University, New York, NY 10027, USA\\[0.25em]%
{jizou@augusta.edu}\quad {szhang2@augusta.edu}\quad {\textsuperscript{}mqiu@augusta.edu}\quad
{cl3607@columbia.edu}
}

\maketitle              % typeset the header of the contribution
\begin{abstract}
Deep learning models are vulnerable to adversarial examples, posing critical security challenges in real-world applications. While Adversarial Training (\textit{AT}) is a widely adopted defense mechanism to enhance robustness, it often incurs a trade-off by degrading performance on unperturbed, natural data. Recent efforts have highlighted that larger models exhibit enhanced robustness over their smaller counterparts. In this paper, we empirically demonstrate that such robustness can be systematically distilled from large teacher models into compact student models.
To achieve better performance, we introduce Dice Adversarial Robustness Distillation (\textit{DARD}), a novel method designed to transfer robustness through a tailored knowledge distillation paradigm. Additionally, we propose \emph{Dice Projected Gradient Descent} (\textit{DPGD}), an adversarial example generalization method optimized for effective attack. Our extensive experiments demonstrate that the \textit{DARD} approach consistently outperforms adversarially trained networks with the same architecture, achieving superior robustness and standard accuracy.
\keywords{ Adversarial examples \and Adversarial attacks  \and Robustness \and Knowledge distillation \and Adversarial training  \and Deep neural networks}
\end{abstract}
\section{Introduction}
\label{sec:introduction}
Deep learning has achieved remarkable success across various domains, from computer vision~\cite{cv} and natural language processing~\cite{nlp} to the control of autonomous systems~\cite{auto}. These advancements are driven by the unparalleled capacity of Deep Neural Networks (\textit{DNNs}) to extract complex features and learn rich representations in diverse real-world applications. Despite their impressive performance, DNNs remain critically susceptible to adversarial examples -- subtle, intentional perturbations that can lead to erroneous predictions while remaining undetected by human perception~\cite{ae}. This vulnerability poses significant security risks in critical applications, such as autonomous driving~\cite{aad}~\cite{aeauto}, large language models~\cite{llm}, and medical diagnosis~\cite{md}, where adversarial attacks can lead to life-threatening outcomes.

Adversarial Training (\textit{AT}) has emerged as one of the most effective approaches for improving robustness by incorporating adversarial examples into the training process~\cite{at}. Despite its efficacy, \textit{AT} suffers from three inherent limitations:
First, \textit{AT}-trained models often exhibit overfitting to specific attack patterns, limiting generalization to novel or adaptive threat vectors.
Second, \textit{AT} frequently leads to degraded performance on clean data (i.e., model's standard accuracy), creating a robustness-accuracy trade-off.
Third, the iterative generation of adversarial examples during training imposes significant computational demands, making \textit{AT} computationally intensive.
These limitations hinder the deployment of robust models in resource-constrained environments.

In response to these challenges, researchers have explored various strategies to improve adversarial robustness. Regularization-based methods, such as adversarial weight perturbation and \textit{TRADES}~\cite{trades}, aim to balance robustness and generalization by imposing constraints on model updates. Data augmentation approaches, leveraging diverse attack strategies or generative models, focus on enhancing defenses against unforeseen adversaries~\cite{da}. Certified defenses offer theoretical robustness guarantees within defined perturbation bound, but often suffer from high computational complexity and limited scalability~\cite{cd}. Despite these efforts, the inherent limitations of adversarial training remain largely unaddressed, highlighting the urgent need for innovative solutions to improve both robustness and generalization.

Recent studies indicate a positive correlation between model size and adversarial robustness, suggesting that larger models tend to exhibit greater resistance to adversarial perturbations compared to smaller counterparts. This is potentially attributed to their capacity for capturing richer feature representations and more effectively mitigating the impact of adversarial noise. Knowledge Distillation (\textit{KD}), a technique for transferring the knowledge of large, high-performing models to smaller, more efficient models, has proven to be effective for maintaining accuracy while reducing computational overhead~\cite{hinton}. Motivated by these insights, we introduce Dice Adversarial Robustness Distillation (\textit{DARD}), a novel framework specifically designed to enhance the adversarial robustness of compact models beyond the capabilities of standard adversarial training.

Our \textit{DARD} framework employs a teacher-student paradigm, where a robustly trained large model guides the learning of a smaller model by distilling knowledge through soft-label probabilities derived from the teacher's output distribution. Unlike traditional hard labels, soft labels retain inter-class relationships, enabling the student model to learn richer and more discriminative representations. Moreover, \textit{DARD} leverages both natural and adversarial soft outputs in a joint objective, carefully balancing the influence of clean and perturbed examples. This balanced approach effectively mitigates the typical degradation of standard accuracy associated with adversarial training while achieving superior adversarial robustness in the student model.

Through extensive experimentation, we demonstrate that our \textit{DARD} framework achieves state-of-the-art robustness-accuracy trade-offs, outperforming adversarially trained baselines of equivalent architecture on the standard datasets. These results highlight the efficacy of knowledge distillation as a powerful mechanism to decouple adversarial robustness from computational overhead, offering valuable insights into the design of deep learning models that are simultaneously robust and computationally efficient. Our key contributions can be summarized as follows:

\renewcommand{\labelitemi}{$\bullet$}
\begin{itemize}[leftmargin=\parindent]
    \item {Dice Adversarial Robustness Distillation (\textit{DARD})}: A novel framework that systematically transfers robustness from large, adversarially trained teachers to compact student models via dual supervision from natural and adversarial soft labels, eliminating the robustness-accuracy trade-off inherent in standard adversarial training.

    \item {Generalization of \textit{DPGD}}: We adapt \textit{Dice Projected Gradient Descent} -- an attack strategy originally designed for semantic segmentation -- to image classification, optimizing its perturbation efficiency and attack success rate through dynamic step-size tuning and channel-wise gradient masking.
    
    \item {Empirical Validation}: Comprehensive experiments show that \textit{DARD}-trained lightweight models (e.g., ResNet-18) achieve better robust accuracy on CIFAR-10 and CIFAR-100, outperforming pre-trained small models of the same architecture, while maintaining near-identical natural accuracy.
\end{itemize}

The rest of this paper is structured as follows. Section~\ref{sec:related-work} reviews related work on adversarial attacks, adversarial training, and knowledge distillation, highlighting key differences between existing approaches and our proposed framework. Section~\ref{sec:proposed-method} introduces \textit{DPGD} adversarial attack and Dice Adversarial Robustness Distillation (\textit{DARD}) framework designed to enhance the student model's acquisition of adversarial features. Section~\ref{sec:experiment} describes the experimental setup, including datasets, evaluation metrics, and implementation details, and presents a comprehensive analysis of the experimental results, demonstrating the effectiveness of \textit{$DARD$} in improving both robust and standard accuracy. Finally, Section~\ref{sec:conclusion} concludes the paper.

\section{Related Work}
\label{sec:related-work}
This section reviews existing research on adversarial attacks, training, and knowledge distillation. This review established the theoretical background and motivation for our proposed Dice Adversarial Robustness Distillation (\textit{DARD}) method.

\subsection{Adversarial Attacks}
\label{subsec:adversarial-attacks}
In 2013, \textit{Szegedy et al.} \cite{ae} first introduced the concept of adversarial examples. Their research demonstrated that subtle, imperceptible to human eyes, can mislead deep neural networks into making incorrect decisions. The process of adding perturbations to the original input is defined as an adversarial attack, and the perturbed inputs are referred to as adversarial examples. Building up this foundation, \textit{Goodfellow et al.}~\cite{fgsm} proposed the Fast Gradient Sign Method (\textit{FGSM}), a groundbreaking approach related to the Gradient Descent algorithm. The goal of generating adversarial examples is to maximize the loss function to induce wrong predictions in the model, as shown in Equation\ref{ae}.

\begin{equation}
    \arg\max_{\|\delta\| \leq \epsilon} \ell(h_{\theta}(x + \delta), y)
\label{ae}
\end{equation}

FGSM efficiently generates adversarial examples by leveraging the gradients of the loss function to the input data, perturbing the inputs in the direction that maximizes the loss, thereby causing the model to make wrong predictions. FGSM is a single-step attack, while unable to achieve optimal adversarial loss. To overcome this limitation, subsequent attacks, such as Projected Gradient Descent (\textit{PGD})~\cite{at}, \textit{DeepFool}~\cite{deepfool} and the Carlini \& Wagner (\textit{CW})~\cite{cw}, these methods adopted multiple-step iterative strategies, gradually approaching more effective adversarial examples through multiple small-step updates. Moreover, for specialized tasks like semantic segmentation, the \textit{SegPGD}~\cite{seg} method distinguished between pixels that are correctly segmented and those that are not in its loss function design, and iteratively adjusts the weighting parameters for these two components to achieve enhanced attack performance. Inspired by these approaches, our proposed Dice Projected Gradient Descent (\textit{PGD}) introduces a Dice Loss coupled with a dynamic weighting strategy to generate even more disruptive adversarial examples in our classification tasks.

\subsection{Adversarial Training}
\label{subsec:adversarial-training}
\textit{Goodfellow et al.}~\cite{fgsm} proposed that deep neural networks are highly susceptible to adversarial examples and adversarial training as a regularization strategy to mitigate overfitting to natural data, similar to the effect of dropout. In their proposed adversarial training method, the loss function was defined as:

\begin{equation}
    J_{\text{adv}}(\theta, x, y) = \alpha J(\theta, x, y) + (1 - \alpha) J(\theta, x_{\text{adv}}, y).
\end{equation}

This approach incorporates both natural and adversarial examples into the training process, thereby enhancing the model's robustness. Generally, $\alpha$ is set to 0.5. Specifically, it employs the \textit{FGSM} to generate adversarial examples. \textit{Madry et al.}~\cite{at} extended this research and introduced a more powerful adversarial training method, namely \textit{PGD} adversarial training. Compared to \textit{FGSM}, \textit{PGD} optimizes adversarial examples through multiple iterative steps, making them more destructive and capable of exploring a broader perturbation space. This advancement significantly improves the robustness of adversarial training, enabling \textit{PGD}-trained models to withstand stronger adversarial attacks while avoiding overfitting to specific perturbation patterns.

\subsection{Knowledge Distillation}

\label{subsec:knowledge-distillation}
Knowledge distillation was originally introduced as a method for model compression and knowledge transfer by \textit{Hinton et al.}~\cite{hinton}. They proposed that a compact student model can be trained to mimic the behavior of a larger, well-trained teacher network. This approach allowed the student model to maintain much of the teacher's predictive capability with lower computational cost. \textit{Romero et al.}~\cite{fitnet} introduced \textit{FitNets}, which leverage intermediate representations, named hints, to facilitate the transfer of knowledge from teacher to student. \textit{Zagoruyko et al.}~\cite{hint} proposed an attention transfer method, wherein the spatial attention maps derived from the teacher network guide the training of the student model, to bolster its representational capability. Researchers integrated adversarial training into the distillation framework to address the adversarial vulnerability in deep neural networks, transferring the robustness from teacher to student~\cite{ard}~\cite{rard}~\cite{rardf}. In these adversarial robustness distillation methods, the student model is trained to learn the teacher's capability on both clean and adversarial examples, enhancing defense against adversarial attacks.

\section{Proposed Method}
\label{sec:proposed-method}
In this section, we introduce our proposed \textit{DARD} method, which integrates our \textit{DPGD} adversarial attack strategy devised into a teacher-student framework to improve adversarial robustness while maintaining high precision on natural inputs. Our \textit{DARD} framework is shown in Fig.\ref{fig2}.
\begin{figure}[t]
\includegraphics[width=\textwidth]{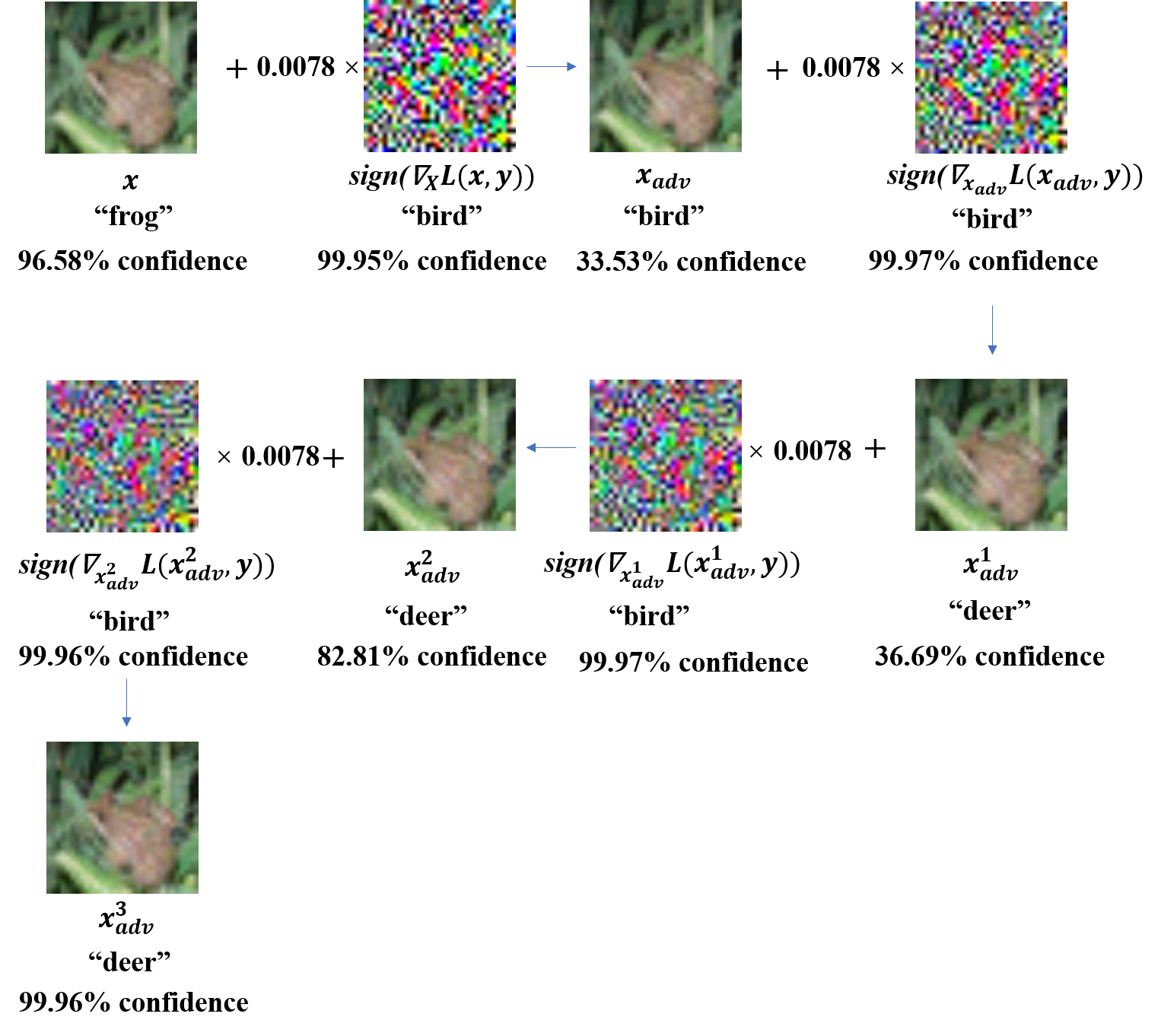}
\caption{Using CIFAR-10 samples and the DPGD algorithm with a step size of $\alpha$ = 2/255, adversarial examples are generated by iteratively adding small perturbations that increase the model’s loss, eventually leading to high-confidence misclassifications.} \label{fig1}
\end{figure}

\subsection{\textit{DPGD}: Dice Projected Gradient Descent}
\label{subsec:dpgd}
\textit{SegPGD}~\cite{seg} was originally proposed for generating adversarial perturbations in semantic segmentation tasks. Its core idea is to use Projected Gradient Descent (PGD) to introduce fine-grained pixel-level perturbations that disrupt the model’s semantic understanding of an image. Inspired by this method, we adapt its iterative perturbation strategy and adversarial loss formulation to the image classification task. 

 Similar to the standard Projected Gradient Descent (\textit{PGD}) framework, \textit{DPGD} seeks to perturb the natural input $X_{nat}$ within an $\ell_\infty$-ball of radius $\epsilon$. However, unlike basic \textit{PGD} attacks, \textit{DPGD} introduces a novel application of Dice Loss and a dynamic weighting scheme that distinguishes between correctly and wrongly predicted samples. \textit{DPGD} can be mathematically defined as follows:

%  Let $X_{nat}$ be a natural example, $y$ its associated ground-truth label, and $S(\cdot;\theta_s)$ the target model parameterized by $\theta_s$. We first define the adversarial example $X_{adv}$ by solving a maximization problem related to the \textit{DPGD} objective:

%  \begin{equation}
% \mathbf{x}_{\text{adv}} \;=\;
% \arg\max_{\|\boldsymbol{\delta}\| \,\le\, \varepsilon}
% \; \mathcal{L}\Bigl(S\bigl(\mathbf{x}_{\text{nat}} + \boldsymbol{\delta}; \theta_s\bigr),\, \mathbf{y}\Bigr).
% \end{equation}

% where $\delta$ is the perturbation constrained by $\|\boldsymbol{\delta}\|_{\infty} \,\le\, \varepsilon$. In typical \textit{PGD} attacks, $L$ is used cross-entropy over all samples in a batch. In our \textit{DPGD}, we replace cross-entropy with a Dice Loss that measures the overlap between the predicted and true label distributions. This choice encourages a larger deviation of the model's predicted probabilities from the correct class, potentially yielding stronger adversarial examples.

\begin{equation}
 \mathbf{X}_{adv}^{\mathrm{t+1}} \;=\; 
\phi^\epsilon\!\Bigl(
  \mathbf{X}_{adv}^{\mathrm{t}} 
  \;+\; 
  \alpha \cdot 
  \mathrm{sign}\!\bigl(
    \nabla_{\mathbf{X}_{adv}^{\mathrm{t}}} 
    L\bigl(f(\mathbf{X}_{adv}^{\mathrm{t}}),  \mathbf{y}\bigr)
  \bigr)
\Bigr),
\end{equation}
where $X^{t}_{adv}$ represents the adversarial example at iteration t, $L\bigl(f(\mathbf{X}_{adv}^{\mathrm{t}}),  \mathbf{y}\bigr)$ denotes the model's output when given the current iteration adversarial input $X_{adv}$, $ \nabla_{\mathbf{X}_{adv}^{\mathrm{t}}} 
L\bigl(f(\mathbf{X}_{adv}^{\mathrm{t}}),  \mathbf{y}\bigr)$ computes the gradient of the loss function to $X^t_{adv}$, $\alpha$ is the step size controlling the magnitude of each update, $sign(\cdot)$ determines the direction of maximum loss increase, $\phi^\epsilon(\cdot)$ is the projection operator that ensures $X^t_{adv}$ remains within an $\ell_\infty$-ball of radius $\epsilon$. The process of \textit{DPGD} is illustrated in Fig.\ref{fig1}.

We define the loss function as two parts, as shown in Equation\ref{eq:loss_function}. The first component calculates the loss for correctly classified samples, while the second component calculates the loss for misclassified samples. To generate highly effective adversarial examples, we introduce dynamically updated weights $\lambda$ and $1-\lambda$. Through this approach, in the early stages of the attack iterations, we primarily perturb correctly classified samples. As the iterations progress, the number of misclassified samples gradually increases. In the later stages of the attack, we focus more on misclassified samples to prevent adversarial examples from becoming ineffective.

\begin{equation}
\mathcal{L} \bigl( f(\mathbf{x}_{\mathrm{adv}}), \mathbf{y} \bigr)
= (1 - \lambda) \mathcal{L}_{p_T} \bigl( \mathbf{x}_{\mathrm{adv}}, \mathbf{y} \bigr)
+ \lambda \mathcal{L}_{\rho_F} \bigl( \mathbf{x}_{\mathrm{adv}}, \mathbf{y} \bigr).
\label{eq:loss_function}
\end{equation}

To ensure that the weight $\lambda$ dynamically adjusts throughout the attack iterations, reflecting different focuses in the early and later stages, we define $\lambda$ as formulated in Equation\ref{eq:weight}. This design allows the attack to prioritize perturbing correctly classified samples in the early iterations while gradually shifting its focus toward misclassified samples in the later iterations, preventing adversarial examples from becoming ineffective. In this equation, t represents the current iteration steps, and T denotes the total iteration steps.

\begin{equation}
\lambda = \frac{{t-1}}{2T}.
\label{eq:weight}
\end{equation}

\subsection{Adversarial Robustness Distillation}
\label{subsec:adversarial-robustness-distillation}
DeepSeek R1 open source enables researchers to distill small models with logical reasoning capabilities at a low cost. Inspired by this approach and supported by previous studies~\cite{mtard}~\cite{ard} indicating that larger models exhibit greater robustness than smaller ones, we explore methods to transfer robustness from large models to compact models effectively. Enhanced robustness can be transferred from a teacher model to a student model, making knowledge distillation a promising approach for improving the robustness of small models. Despite the effectiveness of certain methods in enhancing the robustness of student models, their ability to accurately classify natural images still requires further improvement.

Our proposed method \textit{DARD} aims to not only improve the robustness of the model but also maintain the high accuracy of classifying natural images. Traditional supervised learning typically uses one-hot hard labels to train models. Assuming there is a five-class classification problem, the true label of the input $x$ is a one-hot vector:

\begin{equation}
    y = [0, 1, 0, 0,  0]
\end{equation}
In our \textit{DARD} framework, we replace one-hot hard labels with soft labels, meaning that we use output probabilities as labels instead:
\begin{equation}
    y' = [0.1, 0.7, 0.1, 0.05, 0.05]
\end{equation}

A one-hot hard label only conveys the correct class to the model, whereas a soft label can transfer information about the relationships between different classes. In $DARD$, we leverage not only the probability output vectors of natural images from the large model but also those of adversarial images, using them as labels for adversarial examples in the small model. This design is motivated by the goal of transferring knowledge from the large model to enhance the small model's performance on both standard and adversarial input.
\begin{figure}[t]
\includegraphics[width=\textwidth]{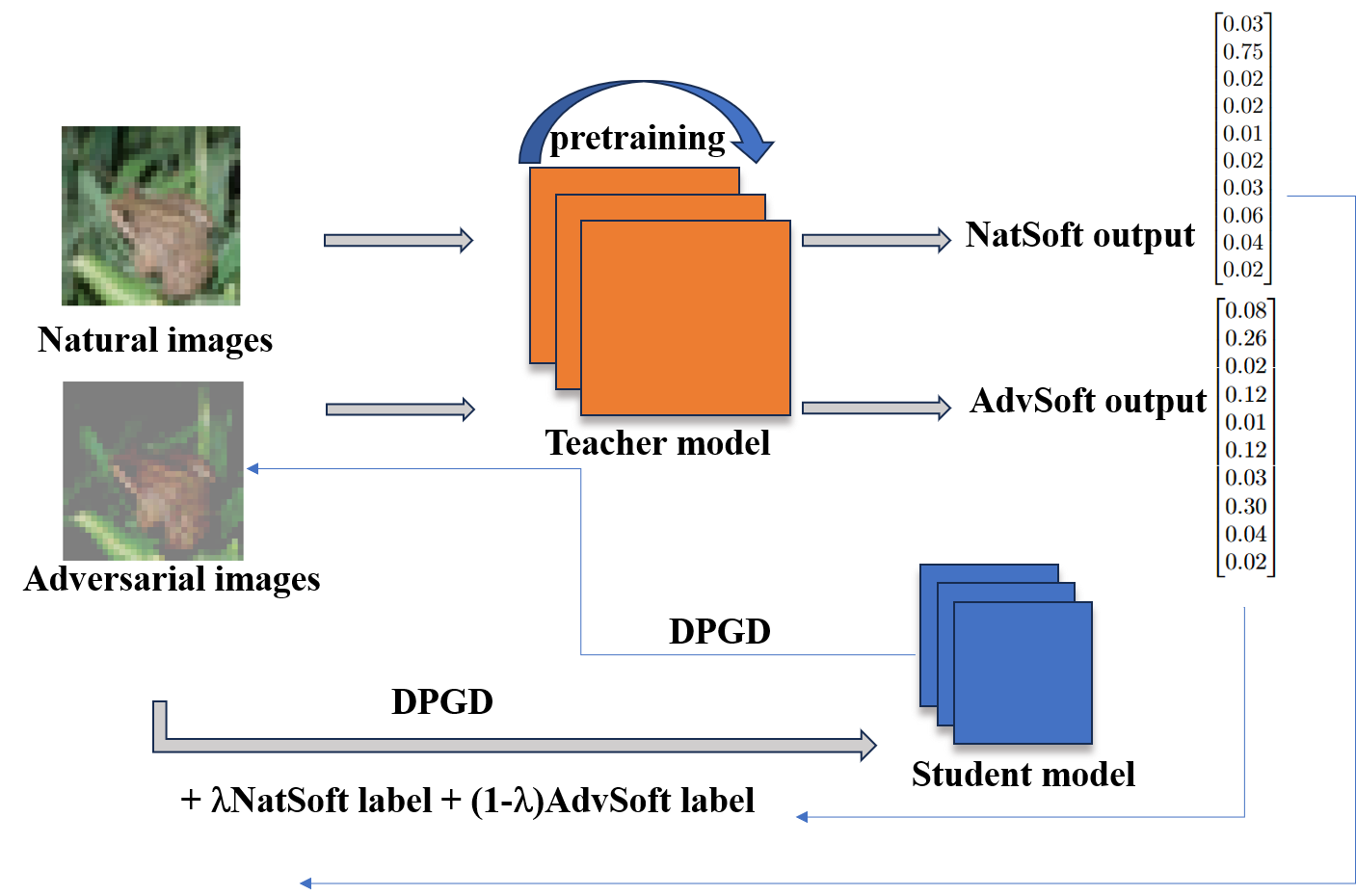}
\caption{The framework of our \textit{DARD} method. First, we generate adversarial examples from the student model using the proposed \textit{DPGD} attack. Next, these adversarial examples are used to train (pretrain) the teacher model adversarially. Finally, we derive a balanced soft probability from the teacher model, combining both natural and adversarial examples, and use this probability as the soft label to train the student model.} \label{fig2}
\end{figure}

Specifically, for both natural examples and adversarial examples, we extract their probability distributions from a large teacher model and jointly use these soft labels to guide the training of a small student model. This approach not only preserves the student model’s accuracy on clean samples but also enables it to leverage the extensive knowledge embedded within the teacher model, thereby enhancing its generalization capabilities. By integrating the teacher model’s insights on natural examples and adversarial perturbations, the student model becomes better equipped to withstand adversarial attacks, ultimately resulting in improved robustness. The student model is trained via an adversarial training framework, wherein adversarial examples are generated using \textit{DPGD} and simultaneously refined through teacher model distillation. The basic optimization framework of $DARD$ can be defined as follows:
\begin{equation}
    \arg\min_{\theta_S} \big((1 - \alpha)\, CE(S(x), y) 
+ \alpha\tau^2 KL(S(x_{\text{adv}}), P(x)\big),
\end{equation}
where $\theta_S$ denotes the student model parameters to be optimized, $S(\cdot)$ is the student model's output, y is the ground-truth label associated with $x$, $CE(S(x_{\text{adv}},y)$ is the cross-entropy loss between the student's prediction on the adversarial examples and the true label, $KL(S(x), P(x))$ denotes the Kullback-Leibler divergence between the student's prediction on the natural input and the teacher model's soft distribution, $P(x)$ is the teacher model's probability distribution on the natural input, and $\alpha$ is the weighting coefficient that balance the distillation loss against the adversarial training loss, $\tau$ is the distillation temperature.

\section{Experiment}
\label{sec:experiment}
\subsection{Experiment setting}
\label{subsec:experiment-setting}
In our experiments, we evaluate the performance of the \textit{DARD} method on two widely used image classification datasets: \textit{CIFAR-10} and \textit{CIFAR-100}~\cite{cifar-10}. We compare our method with natural training and two state-of-the-art adversarial robustness methods: Standard Adversarial Training (\textit{SAT})~\cite{at} and Adversarially Robustness Distillation (\textit{ARD})~\cite{ard}. All experiments run on CloudLab using its c240g5 node, each equipped with two Intel Xeon Silver 4114 CPUs (10 cores each, 2.20 GHz), 192 GB of ECC DDR4-2666 memory, an NVIDIA P100 GPU (12 GB), a 1 TB 7200 RPM SAS hard drive, and a 480 GB Intel DC S3500 SSD.

\subsubsection{Student and Teacher Networks.}
\label{subsubsec:student}
In our experiments, we consider \textit{ResNet-18}~\cite{resnet} as the student model, \textit{ResNet-56}~\cite{resnet} as the teacher model following previous work. The teacher model is pre-trained before adversarial robustness distillation. 

\begin{table}[ht]
    \centering
    \normalsize
    \renewcommand{\arraystretch}{1.2}
    % \resizebox{\textwidth}{!}{%
    \begin{tabular*}{\textwidth}{@{\extracolsep{\fill}\quad}lcccccc}
        \toprule
        Defense 
        & Clean & FGSM & PGD20 & T-PGD & BIM & AutoAttack  \\
        \midrule
        Natural  
        & \textbf{93.49}\% & 17.03\% & 0.00\% & 0.55\% & 0.00\% & 0.00\% \\
        SAT & 81.33\% & 55.62\% & 49.68\%  & 78.42\% & 50.27\% & 46.63\% \\
        ARD & 84.01\% & 57.40\% & 50.71\% & 76.42\% & 50.12\% & 45.78\% \\
         \textbf{TDARD} & 85.88\% & 54.78\% & 44.35\% & \textbf{82.55}\% & 44.12\% & 42.58\% \\
        \textbf{DARD} & 83.56\% & \textbf{57.53}\% & \textbf{52.63}\% & 81.11\% & \textbf{53.07}\% & \textbf{47.75}\% \\
        \bottomrule
    \end{tabular*}%
    % }
    \vspace{1ex}
    \caption{White-box robustness of ResNet-18 on CIFAR-10}
    
    \label{tab:D1}
    
\end{table}

\begin{table}[ht]
    \centering
    \normalsize
    \renewcommand{\arraystretch}{1.2}
    % \resizebox{\textwidth}{!}{%
    \begin{tabular*}{\textwidth}{@{\extracolsep{\fill}\quad}lcccccc}
        \toprule
        Defense 
        & Clean & FGSM & PGD20 & T-PGD & BIM & AutoAttack  \\
        \midrule
        Natural  
        & \textbf{75.01}\% & 7.05\% & 0.00\% & 0.26\% & 0.00\% & 0.00\% \\
        SAT & 58.59\% & 28.18\% & 23.79\%  & 55.64\% & 23.80\% & 20.85\% \\
        ARD & 59.50\% & 29.30\% & 25.61\% & 53.64\% & 21.60\% & 21.54\% \\
         \textbf{TDARD} & 65.06\% & 29.90\% & 24.33\% & \textbf{57.96}\% & 23.80\% & 19.60\% \\
        \textbf{DARD} & 61.13\% & \textbf{32.42}\% & \textbf{28.80}\% & 57.40\% & \textbf{28.74}\% & \textbf{23.44}\% \\
        \bottomrule
    \end{tabular*}%
    % }
    \vspace{1ex}
    \caption{White-box robustness of ResNet-18 on CIFAR-100}
    \vspace{-6ex}
    \label{tab:D2}
\end{table}
\subsubsection{Adversarial Attacks.}
\label{subsubsec:adversarial-attacks}
To comprehensively evaluate the robustness of our proposed method, we compare it against five widely used adversarial attack methods: \textit{FGSM} (Fast Gradient Sign Method), \textit{PGD}
 (Projected Gradient Descent), \textit{T-PGD} (Target Projected Gradient Descent)~\cite{t-pgd}, \textit{BIM} (Basic Iterative Method)~\cite{bim}, and \textit{AutoAttack}~\cite{autoattack}. These attack methods cover a range of adversarial attack strategies, including single-step attacks (\textit{FGSM}) and multi-step iterative attacks (\textit{PGD}, \textit{T-PGD}, \textit{BIM}, and \textit{AutoAttack}), ensuring diversity in adversarial examples. \textit{AutoAttack} acts as a comprehensive attack framework that combines multiple adversarial strategies, including \textit{Auto-PGD}, \textit{APGD-CE}, \textit{PGD-DLR}, and \textit{Square Attack}. As a popular strategy for rigorously testing model robustness, \textit{AutoAttack} subjects defense mechanisms to a broad and stringent evaluation. In our experiments, we standardized the hyper parameters for all multi-step iterative adversarial attacks to ensure a fair comparison. Specifically, we set the number of iterations to 20, the maximum perturbations value $\epsilon$ = 8/255, and the step size $\alpha$ = 2/255. These setting ensure a controlled and consistent evaluation of the model's robustness against iterative adversarial attacks.

\begin{table}[ht]
    \centering
    \normalsize
    % \renewcommand{\arraystretch}{1.2} 
    % \resizebox{\textwidth}{!}{
    \begin{tabular*}{\textwidth}{@{\extracolsep{\fill}\quad}llccc}
    \toprule
    \textbf{Attack} & \textbf{Defense} & \textbf{Clean} & \textbf{Robust} & \textbf{W-Robust}\\
    \midrule
    \multirow{6}{*}{FGSM} 
        & Natural & 93.49\% & 17.03\% & 55.26\% \\
        & SAT     & 81.33\% & 55.62\% & 68.48\% \\
        & ARD     & 84.01\% & 57.40\% & 70.71\% \\
        & TDARD   & 85.88\% & 54.78\% & \textbf{70.33\%} \\
        & DARD    & 83.56\% & 57.53\% & \textbf{70.55\%} \\
    \midrule
    \multirow{6}{*}{PGD20} 
        & Natural & 93.49\% & 0.00\% & 46.75\% \\
        & SAT     & 81.33\% & 49.68\% & 65.51\% \\
        & ARD     & 84.01\% & 50.71\% & 67.36\% \\
        & TDARD   & 85.88\% & 44.35\% & \textbf{65.12\%} \\
        & DARD    & 83.56\% & 52.63\% & \textbf{68.10\%} \\
    \midrule
    \multirow{6}{*}{T-PGD} 
        & Natural & 93.49\% & 0.55\% & 47.02\% \\
        & SAT     & 81.33\% & 78.42\% & 79.88\% \\
        & ARD     & 84.01\% & 76.42\% & 80.22\% \\
        & TDARD   & 85.88\% & 82.55\% & \textbf{84.22\%} \\
        & DARD    & 83.56\% & 81.11\% & \textbf{82.34\%} \\
    \midrule
    \multirow{6}{*}{BIM} 
        & Natural & 93.49\% & 0.00\% & 46.75\% \\
        & SAT     & 81.33\% & 50.27\% & 65.80\% \\
        & ARD     & 84.01\% & 50.12\% & 67.07\% \\
        & TDARD   & 85.88\% & 44.12\% & \textbf{65.00\%} \\
        & DARD    & 83.56\% & 53.07\% & \textbf{68.32\%} \\
    \midrule
    \multirow{6}{*}{AutoAttack} 
        & Natural & 93.49\% & 0.00\% & 46.75\% \\
        & SAT     & 81.33\% & 46.63\% & 63.98\% \\
        & ARD     & 84.01\% & 45.78\% & 64.90\% \\
        & TDARD   & 85.88\% & 42.58\% & \textbf{64.23\%} \\
        & DARD    & 83.56\% & 47.75\% & \textbf{65.66\%} \\
    \bottomrule
    \end{tabular*}
    % }
    \vspace{1ex}
    \caption{White-box trade-off robustness of ResNet-18 on CIFAR-10}
    \vspace{-3ex}
    \label{tab:DW}
\end{table}

\subsection{Training and Evaluation}
\label{subsec:training-and-evaluation}
We employ the Stochastic Gradient Descent (\textit{SGD}) optimizer to train our student model, setting the learning rate to 0.1, momentum to 0.9, and weight decay to 5e-4. This configuration is commonly used in training neural networks as it effectively balances convergence speed and generalization. In our \textit{DARD} method, we enhance the adversarial robustness of the student model by adopting an adversarial strategy that employs only adversarial examples. Specifically, for each original sample, we generate its corresponding adversarial example by \textit{DPGD} and use these adversarial examples exclusively to train the student model. Meanwhile, we evaluated various values for the hyperparameter $\lambda$ during the construction of soft labels for the student model and found that setting $\lambda$ to 0.5 yielded the preferred performance. In other words, for a given sample, we obtain the teacher model's prediction distribution on the natural sample and on the adversarial example, then average these two distributions to form the soft label. This allows the student model to simultaneously absorb the teacher's knowledge of both clean and adversarial distributions, thereby preserving an understanding of the clean data distribution despite training solely on adversarial examples and ultimately improving overall robustness and generalization. In our experiments, we also evaluated another adversarial training strategy within the \textit{DARD} framework, referred to as \textit{TDARD}, which trains the student model using both natural and adversarial examples simultaneously. Although both strategies operate under the same framework, they emphasize different aspects. Our experimental results indicate that the first strategy demonstrates a more comprehensive performance. The results are presented in Table\ref{tab:D1} and \ref{tab:D2}.

In our comparative experiments, to ensure consistency with \textit{DARD} approach, we adopted the core design principles of the \textit{SAT} and \textit{ARD} methods and implemented them on the same \textit{ResNet18} architecture used by our student model. Additionally, to better compare these methods and \textit{DARD} balance accuracy on both clean and adversarial examples, we introduce Weighted Robust Accuracy~\cite{wra} as our evaluation metric. Specifically, this metric takes into account the accuracies on clean and adversarial examples by averaging them, thus providing a comprehensive reflection of the model's performance under both natural and adversarial examples. Table\ref{tab:DW} shows the results.

\subsection{Ablation Study}
\label{subsec:ablation-study}
\begin{figure}[t]
 \includegraphics[width=\textwidth]{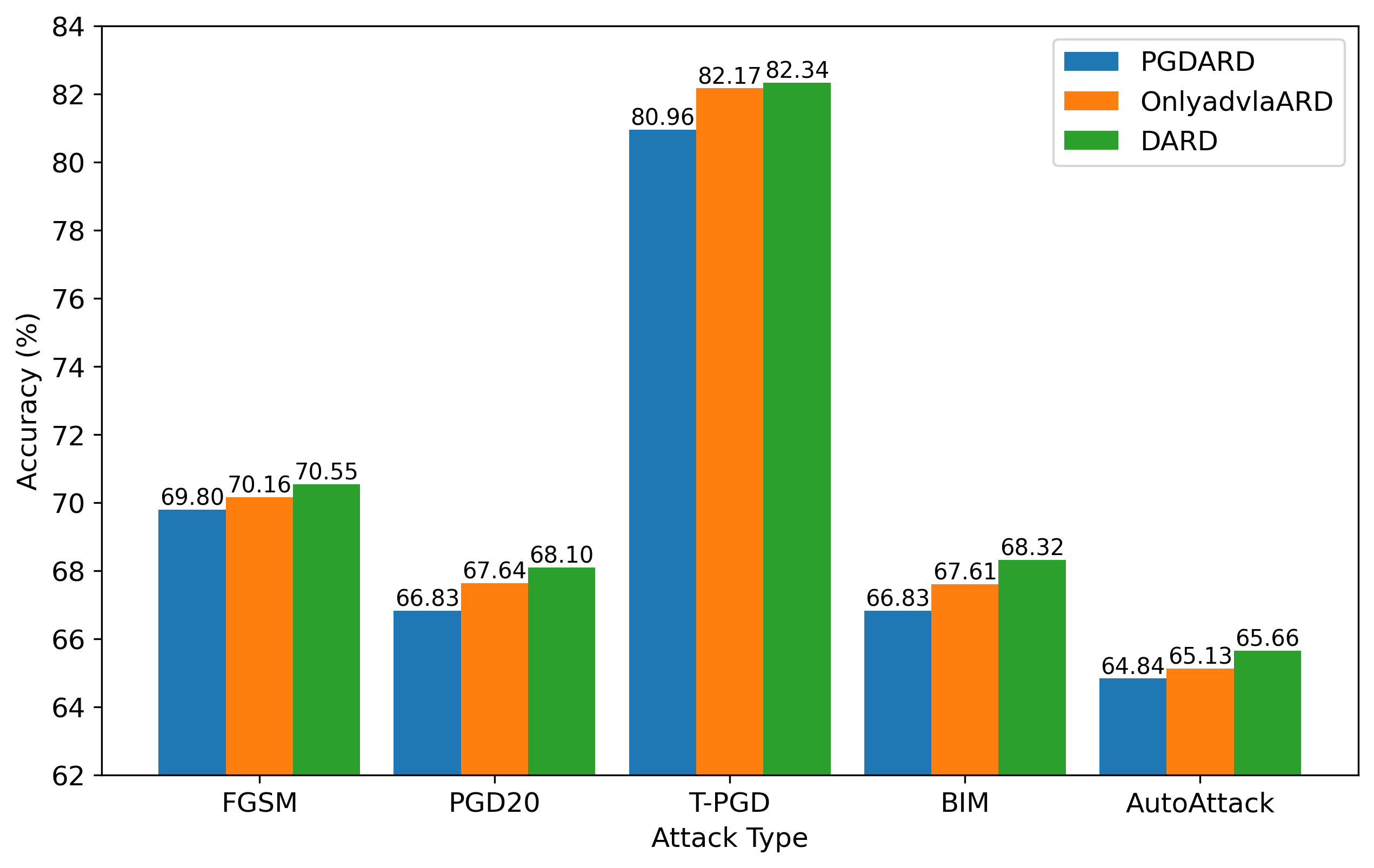}
 \caption{Based on ablation experiments using the ResNet-18 model, we compare the performance differences among three adversarial robustness distillation strategies. Specifically, the PGDARD method employs the conventional PGD algorithm to generate adversarial examples, which are then used for adversarial knowledge distillation. In contrast, the OnlyadvlaARD method solely utilizes the teacher model’s soft outputs on adversarial examples as guidance for the student model. Our proposed \textit{DARD} approach, on the other hand, leverages the \textit{DPGD} algorithm to generate adversarial examples and integrates the teacher model’s soft outputs from both clean and adversarial examples during the distillation process.} \label{abla}
\end{figure}

\begin{figure}[t]
\centering
 \includegraphics[width=0.8\textwidth]{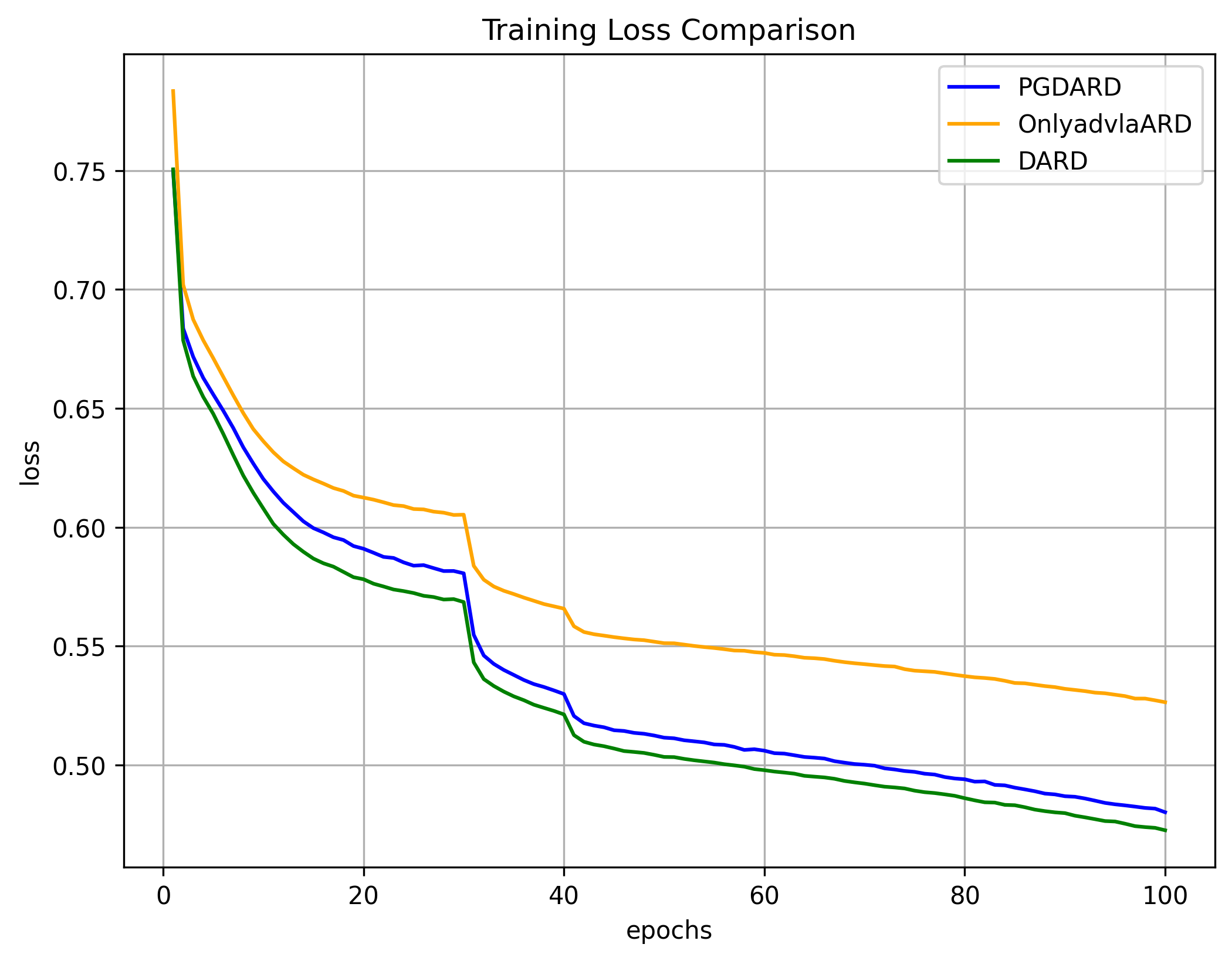}
 \caption{The training loss with ResNet-18 student network distilled using PGDARD, OnlyadvlaARD, and DARD on CIFAR-10.} \label{loss}
\end{figure}

To gain deeper insights into the role of each part in the \textit{DARD} framework and their contributions to adversarial robustness distillation, we conducted a series of systematic ablation studies. The results are shown in Fig.\ref{abla}, and the change of total loss is shown in Fig.\ref{loss}.

\subsubsection{Effect of the \textit{DPGD} Module.} 
\label{subsubsec:effect}
Specifically, we replaced our proposed \textit{DPGD} with the traditional 20-step \textit{PGD} to quantitatively assess the impact of \textit{DPGD} on the overall adversarial robustness and classification performance of \textit{DARD}. The results highlight the effectiveness of the \textit{DPGD} module within the \textit{DARD} framework. Specifically, our \textit{DARD} method, enhanced with \textit{DPGD}, achieves a more balanced performance across both natural and adversarial examples.
\subsubsection{Effect of Combining Clean and Adversarial Soft Label.}
\label{subsubsec:combining}
In addition, during adversarial robustness distillation, we used only the soft output of the teacher model on adversarial examples as soft labels for the student model to investigate whether incorporating the teacher model's soft output on both natural and adversarial examples is necessary. The ablation experimental results demonstrate that the \textit{DARD} method leverages the guidance of the teacher model on both natural and adversarial examples, enabling it to capture the teacher model's behavior more comprehensively. As a result, \textit{DARD} achieves superior and more balanced performance across both natural and adversarial examples.

Through these two ablation studies, we obtained a more comprehensive understanding of the crucial role played by \textit{DPGD} in \textit{DARD} and verified the effectiveness of incorporating the soft outputs of both natural and adversarial examples in enhancing model robustness. Based on the results shown in Fig.~\ref{abla} and~\ref{loss}, we conclude that \textit{DARD} exhibits optimal performance in terms of both training convergence and adversarial robustness across all ablation studies.

\section{Conclusion}
\label{sec:conclusion}
In this work, we presented \textit{Dice Adversarial Robustness Distillation} (\textit{DARD}), a novel knowledge distillation framework that simultaneously enhances the adversarial robustness of deep learning models and maintains high natural accuracy on clean data. A key insight from our study within \textit{DARD} is the surprising effectiveness of adversarial-example-only training in achieving a superior equilibrium between standard and robust accuracy, providing a promising perspective for further research aimed at optimizing model performance across diverse operational conditions. 

Our extensive empirical evaluations on \textit{CIFAR-10} and \textit{CIFAR-100} datasets demonstrated that \textit{DARD} significantly outperforms traditional adversarial training baselines, including \textit{SAT} and \textit{ARD}, in various attack scenarios. These results underscore the efficacy of our approach in addressing the robustness-accuracy trade-off. Nevertheless, \textit{DARD} does exhibit certain limitations, notably the computational demands associated with knowledge distillation and adversarial sample generation. 
Future research will focus on improving the efficiency of the distillation process and minimizing computational overhead without compromising robustness. Additionally, we plan to extend the applicability of the \textit{DARD} framework by evaluating its performance on larger-scale datasets and in real-world adversarial settings, ensuring its relevance and efficacy across a broader range of applications. Moreover, exploring robustness against emerging stealthy threats such as clean-label backdoor attacks~\cite{Naback} is a critical direction for future research, highlighting the importance of developing more comprehensive defensive framework.


\end{document}